\documentclass[10pt,twocolumn,letterpaper]{article}

\usepackage[pagenumbers]{iccv} 
\usepackage{multirow, booktabs}
\usepackage[accsupp]{axessibility}
\definecolor{iccvblue}{rgb}{0.21,0.49,0.74}
\usepackage[pagebackref,breaklinks,colorlinks,allcolors=iccvblue]{hyperref}

\title{MUSE: Multi-Subject Unified Synthesis via Explicit Layout Semantic Expansion}

\author{
Fei Peng$^{1,2}$\thanks{Equal contribution.\quad$^\dagger$Corresponding author.} \quad 
Junqiang Wu$^2\footnotemark[1]$ \quad 
Yan Li$^2\footnotemark[2]$ \quad % 分配新标记
Tingting Gao$^2$ \quad 
Di Zhang$^2$ \quad 
Huiyuan Fu$^1$\footnotemark[2]\\
$^1$Beijing University of Posts and Telecommunications, China\quad
$^2$Kuaishou Technology
}

\begin{document}
\maketitle
\begin{abstract}

Existing text-to-image diffusion models have demonstrated remarkable capabilities in generating high-quality images guided by textual prompts. However, achieving multi-subject compositional synthesis with precise spatial control remains a significant challenge. In this work, we address the task of layout-controllable multi-subject synthesis (LMS), which requires both faithful reconstruction of reference subjects and their accurate placement in specified regions within a unified image. While recent advancements have separately improved layout control and subject synthesis, existing approaches struggle to simultaneously satisfy the dual requirements of spatial precision and identity preservation in this composite task. To bridge this gap, we propose MUSE, a unified synthesis framework that employs concatenated cross-attention (CCA) to seamlessly integrate layout specifications with textual guidance through explicit semantic space expansion. The proposed CCA mechanism enables bidirectional modality alignment between spatial constraints and textual descriptions without interference. Furthermore, we design a progressive two-stage training strategy that decomposes the LMS task into learnable sub-objectives for effective optimization. Extensive experiments demonstrate that MUSE achieves zero-shot end-to-end generation with superior spatial accuracy and identity consistency compared to existing solutions, advancing the frontier of controllable image synthesis. Our code and model are available at \href{https://github.com/pf0607/MUSE}{https://github.com/pf0607/MUSE}.

\end{abstract}    
\vspace{-3mm}
\section{Introduction}
\label{sec:intro}

Recent advances in text-to-image generation \cite{podell2023sdxl, ramesh2022hierarchical, rombach2022high, gu2022vector, saharia2022photorealistic, dhariwal2021diffusion,chang2023muse,razzhigaev2023kandinsky, betker2023improving,chen2023pixart, esser2024scaling} built upon diffusion models \cite{ dhariwal2021diffusion, song2020score,ho2020denoising, song2020denoising} have demonstrated unprecedented capabilities in synthesizing photorealistic images from textual descriptions. The emergence of architectural innovations like Stable Diffusion \cite{rombach2022high} has particularly enabled efficient high-resolution generation while maintaining strong text-image alignment. However, achieving precise spatial control over multiple visual elements remains an open challenge, as current models often struggle to faithfully reconstruct specific subjects and position them in designated image regions through textual prompts alone.

This limitation motivates our investigation of layout-controllable multi-subject synthesis (LMS) - a critical task requiring simultaneous satisfaction of two objectives: 1) accurate reconstruction of multiple subjects with identity preservation, and 2) strict adherence to user-specified spatial arrangements within a unified image. The LMS task holds significant practical value for applications ranging from graphic design to interactive storytelling, where precise control over both visual content and composition is essential. Current approaches addressing these requirements can be categorized into two distinct research directions, each with inherent limitations when applied to LMS.

Existing layout control methods primarily adopt two paradigms. Training-free techniques \cite{bar2023multidiffusion, kim2023dense, chefer2023attend, yang2023reco, feng2022training, chen2024training, xie2023boxdiff} achieve spatial control through post-hoc fusion of independently generated instances during inference, but suffer from inconsistent lighting and perspective across regions. Condition-driven approaches \cite{gafni2022make,avrahami2023spatext,zhang2023adding,chen2024anydoor,li2023gligen, wang2024instancediffusion, zhou2024migc, wu2024ifadapter, huang2024realcustom} incorporate spatial constraints via additional control modules during training, yet typically lack explicit mechanisms for reference-based subject synthesis. Conversely, subject-driven generation methods \cite{ruiz2023dreambooth,gal2022image,qiu2023controlling,koh2024generating,ye2023ip,wang2024instantid,zhang2024ssr, patel2024lambda, ma2024subject,sun2024generative} excel at reconstructing individual or multiple subjects through textual inversion or adapter networks, but fail to scale effectively to multi-subject scenarios with precise layout requirements. Recent hybrid attempts \cite{wang2024ms,li2023gligen} to combine these capabilities reveal fundamental architectural limitations: parameter-intensive fusion mechanisms \cite{li2023gligen} become computationally prohibitive at scale, while decoupled cross-attention approaches \cite{wang2024ms} exhibit control collision between layout and text conditions.

To overcome these limitations, we present MUSE (Multi-Subject Unified Synthesis via Explicit Layout Semantic Expansion), a novel framework that achieves precise layout control and identity-preserving subject synthesis through three key innovations. First, we introduce concatenated cross-attention (CCA), an explicit semantic space expansion mechanism that enables bidirectional alignment between layout specifications and textual guidance without interference. Second, we develop a progressive two-stage training strategy that decouples layout control learning from subject synthesis optimization, effectively mitigating control collision during joint training. Third, our architecture requires no test-time fine-tuning while supporting zero-shot generation of complex multi-subject compositions.

Our primary contributions are threefold:
1) We identify key technical barriers preventing existing methods from achieving layout-controllable multi-subject synthesis (LMS) and propose an efficient LMS framework—MUSE.
2) We propose an explicit layout semantic expansion method that leverages the newly introduced concatenated cross-attention to enable compatibility between layout control and text control, achieving more precise and effective layout-controllable image generation.
3) We devise a progressive training paradigm that decomposes the LMS objective into sequential layout control acquisition and subject synthesis refinement phases, effectively resolving optimization conflicts inherent in joint training.
% \end{itemize}

Extensive quantitative evaluations and ablation studies demonstrate that MUSE achieves state-of-the-art performance in both layout accuracy and subject fidelity, while maintaining efficient generation speeds. Our framework establishes new capabilities for controllable image synthesis, significantly advancing the practical deployment of diffusion models in professional design workflows.

\vspace{-2mm}

\section{Related Work}
\label{sec:related work}

\noindent \textbf{Text-to-Image Generation}
\cite{balaji2022ediff, nichol2021glide, podell2023sdxl, ramesh2022hierarchical, rombach2022high, gu2022vector, saharia2022photorealistic, dhariwal2021diffusion, song2020score,chang2023muse,ho2020denoising, song2020denoising, brock2018large,goodfellow2020generative,reed2016generative,razzhigaev2023kandinsky, betker2023improving,chen2023pixart, esser2024scaling}  utilize text as control conditions to generate images that meet specified requirements.
 In recent years, diffusion models \cite{ dhariwal2021diffusion, song2020score,ho2020denoising, song2020denoising} have demonstrated impressive performance in image generation, but their efficiency is compromised due to the necessity of multiple steps of noise prediction in pixel space. Latent Diffusion Model (LDM) \cite{rombach2022high} improves this issue by integrating Variational Auto-Encoder (VAE) \cite{kingma2013auto} technique to conduct the denoising process in a compressed latent space, significantly enhancing the efficiency of image diffusion models. 

\noindent \textbf{Layout Control for Image Generation} 
serves as a complement to the text control by providing additional layout conditions, such as segmentation masks and bounding boxes. Training-free methods \cite{bar2023multidiffusion, kim2023dense, chefer2023attend, yang2023reco, feng2022training, chen2024training, xie2023boxdiff} introduce layout guidance operations during inference on pre-trained generation models. For example, MultiDiffusion \cite{bar2023multidiffusion}, NoiseCollage \cite{shirakawa2024noisecollage} and SceneDiffusion \cite{ren2024move} denoise each instance separately and then perform a fusion of the multiple instances.   
Additionally, some methods \cite{gafni2022make,avrahami2023spatext,zhang2023adding,chen2024anydoor,li2023gligen, wang2024instancediffusion, zhou2024migc, wu2024ifadapter, huang2024realcustom} fine-tune pre-trained models to accept layout control conditions. Make-a-scene \cite{gafni2022make}, Spa-text \cite{avrahami2023spatext}, ControlNet \cite{zhang2023adding} and Anydoor \cite{chen2024anydoor} utilize segmentation masks to generate images that conform to the shapes defined by these masks. GLIGEN \cite{li2023gligen}, InstanceDiffusion \cite{wang2024instancediffusion}, MIGC \cite{zhou2024migc, zhou2024migc++} and IFAdapter \cite{wu2024ifadapter} allow for more flexible layout control by accepting various layout conditions, such as bounding boxes, points, and masks.

\noindent \textbf{Subject Synthesis} has made significant progress in single-subject methods \cite{ruiz2023dreambooth,gal2022image,han2023svdiff,qiu2023controlling,koh2024generating,ye2023ip,wang2024instantid,wei2023elite,li2023blip}. Dreambooth \cite{ruiz2023dreambooth} binds reference images to specific labels and fine-tunes the U-Net. 
IP-Adapter \cite{ye2023ip} utilizes CLIP \cite{radford2021learning} to encode reference images and trains additional decoupled attention networks to inject image control. InstantID \cite{wang2024instantid} enhances facial personalization by incorporating an additional face encoder. Real-Custom \cite{huang2024realcustom} constructs an adaptive module to select appropriate encoded image features.
Multi-subject synthesis has attracted increasing research interest.
Custom Diffusion \cite{kumari2023multi} reduces the parameter update requirements of DreamBooth, allowing multiple subjects to be bound to different labels simultaneously. Mix-of-Show \cite{gu2023mix} binds multiple subjects by additionally training LoRA. Zero-shot methods \cite{zhang2024ssr, wu2024multigen, patel2024lambda, wang2024ms, ma2024subject, pan2023kosmos,sun2024generative} are simpler and more practical for real-world applications.
SSRencoder \cite{zhang2024ssr} extracts subject features from images by relating them to the corresponding text in the prompt. $\lambda$-eclipse \cite{patel2024lambda} maps encoded reference image and text features into the same representation in the latent space.
MultiGen \cite{wu2024multigen} fine-tunes the pre-trained generative model to incorporate multi-modal control. 
MS-Diffusion \cite{wang2024ms} and SubjectDiffusion \cite{ma2024subject} fuse the subject images and coordinates through a network and then inject them via adapters. 
\section{Method}
\label{sec:method}

\subsection{Preliminaries}

Stable Diffusion \cite{rombach2022high} is a commonly used text-to-image diffusion model that performs diffusion operations in a compressed latent space to improve the efficiency of high-resolution image generation. In UNet, cross-attention (CA) layers are designed specifically to receive control conditions. Formally, let $Q$ denote the query obtained by projecting the unfolded latent space image features via a linear layer, and  $K$, \( V \) represent the key and value derived by projecting the features encoded from the control text by CLIP \cite{radford2021learning}. The token dimension of \( K \) is \( d\). $\sigma$ denotes the softmax operation. The CA is calculated as follows:
\begin{equation}
\label{eq:cross attention}
\text{CA} = \sigma \left(\frac{QK^T}{\sqrt{d}}\right)V.
\end{equation}
% \vspace{0.5mm}
Decoupled cross-attention (DCA) is a commonly used method for plugin-based injection of additional control conditions. It operates by freezing the pre-trained UNet and training additional linear layers within the original CA layer to project encoded additional control features into key and value. Then compute new CA with the pre-trained query and add the result to original text-driven CA result. For instance, in a subject synthesis task, the calculation of DCA can be formulated as:
\begin{equation}
\label{eq:decoupled cross attention}
\text{DCA}= \sigma\left(\frac{Q K^T}{\sqrt{d}}\right)V +\lambda \cdot \sigma\left(\frac{Q K_I^T}{\sqrt{d}}\right)V_I.
\end{equation}
\( K_I \) and \( V_I \) are the key and value projected from the image features. \( \lambda \) is the adjustable control strength scale, typically taking values between 0 and 1. DCA effectively leverages the prior knowledge of the pre-trained model, allowing for efficient training with minimal increase in parameter count.

\subsection{Overview}

\textbf{Task Definition.} In this paper, we introduce the layout-controllable multi-subject synthesis (\textbf{LMS}) task. We define the input control conditions as follows: the global prompt text, along with additional inputs for \( N \) subjects, which include each subject’s reference image, bounding box coordinates, and class text. These inputs are denoted as \( T_{\text{global}} \), \( I_i \), \( B_i \), and \( T_i \) for \( 1 \leq i \leq N \), respectively. The image generation model is required to produce an image that not only aligns with \( T_{\text{global}} \) but also ensures that each subject appears in its specified \( B_i \) and resembles the corresponding \( I_i \).

\noindent \textbf{Limitations.} Current methods that incorporate these conditions simultaneously have notable limitations. GLIGEN \cite{li2023gligen} adds new gated self-attention (SA) layers between the SA and CA layers of the UNet to accept control inputs, which significantly increases the size of the diffusion model (\eg, SDXL grows from 2.56B to 4.3B), making it overly burdensome and difficult to train as current model sizes continue to increase.
Existing commonly used control methods \cite{ye2023ip,zhou2024migc,zhou2024migc++,wang2024ms,zhang2024ssr,wang2024instantid,wu2024ifadapter} typically use lighter-weight DCA method that is compatible with pre-trained models and has been successfully applied in larger models like SDXL \cite{podell2023sdxl}. MS-Diffusion \cite{wang2024ms} uses DCA on the SDXL model to explore LMS tasks and demonstrates some effectiveness. However, we observe that its layout control capability is relatively poor. This limitation arises partly due to the control collision issue associated with the DCA method, as discussed in \cref{subsec:explicit layout semantic expansion}, and partly because layout control and multi-subject synthesis are two challenging sub-tasks to reconcile effectively during training.

\noindent \textbf{Proposed method.} To enable effective LMS, we propose the MUSE framework. We decompose the task into two progressive stages. First, we propose an explicit layout semantic expansion method that enables compatibility between layout and text controls, achieving text-aligned, layout-controllable image generation. Second, we further train the subject synthesis DCA network based on the pre-trained layout-controllable generation model. With the combination of the advantages from the two progressive stages, we ultimately obtain an effective zero-shot, end-to-end LMS model.

\subsection{Explicit Layout Semantic Expansion for Text}
\label{subsec:explicit layout semantic expansion}
\begin{figure}[t]
  \centering
   \includegraphics[width=0.9\linewidth]{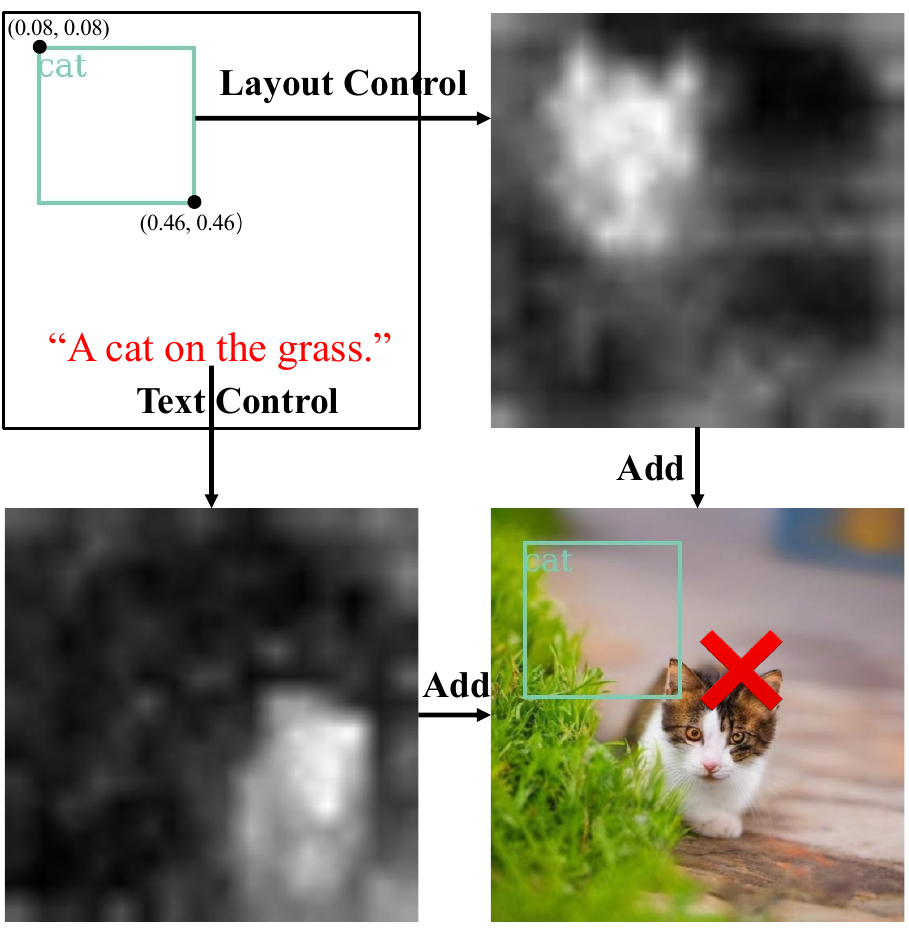}
   \vspace{-2mm}
   \caption{The control collision issue. This occurs when the attention generated by layout control conflicts with that of text control over specific layout regions, ultimately leading to generated images that fail to meet layout control conditions.
}
   \vspace{-3.5mm}
   \label{fig:Control Collision}
\end{figure}
For the standalone layout control task, subject synthesis is not required. Instead, only the additional subject text \( T_i \) and subject position \( B_i \) are considered. We follow existing methods \cite{li2023gligen, wang2024instancediffusion, zhou2024migc, wang2024ms, wu2024ifadapter}, which typically fuse the encoded features of \( T_i \) with the encoded \( B_i \) to obtain the final layout control representation, known as the grounding token and denoted as \( G_T^i \). This is calculated as follows:
\begin{equation}
\label{eq:grounding text}
\text{\( G_T^i \)} = \text{MLP}([f_T^i, \text{Fourier}(B_i)]),  \quad(1\leq i \leq N)
\end{equation}
where \( B_i \) is Fourier encoded and \( f_T^i \) is the encoded feature of \( T_i \), often the class token extracted by CLIP \cite{radford2021learning}. $[\cdot]$ denotes the concatenation operation. Then the concatenated features are fused through a Multi-Layer Perceptron network (MLP). To achieve multi-subject layout control, it is sufficient to concatenate the grounding tokens of all subjects, represented as \( G_T = [G_T^1, G_T^2, \dots, G_T^N] \).

\noindent \textbf{Control Collision.} Compared to some methods \cite{wang2024instancediffusion,li2023gligen} that incorporate large gated SA layers to receive the $G_T$, other methods \cite{zhou2024migc,wang2024ms,wu2024ifadapter} using DCA offer higher training and inference efficiency, achieving significant progress in layout control. However, layout control with DCA presents an issue we term control collision: without injected layout control, image generation under a text prompt generally results in random layouts that may be aligned with the broader distribution of training data. When injected layout control information diverges from this data distribution, as shown in \cref{fig:Control Collision}, two distinct layouts may emerge from the CA heatmaps of text and layout control, causing collision when combined through element-wise addition in \cref{eq:decoupled cross attention}. This results in poor layout control in generated images. Methods like MIGC \cite{zhou2024migc}, NoiseCollage \cite{shirakawa2024noisecollage}, and SceneDiffusion \cite{ren2024move} try to solve this by separately generating global backgrounds and individual instances before merging, but at a high computational cost, potentially leading to inconsistencies among instances.
\begin{figure}[t]
  \centering
   \includegraphics[width=1.0\linewidth]{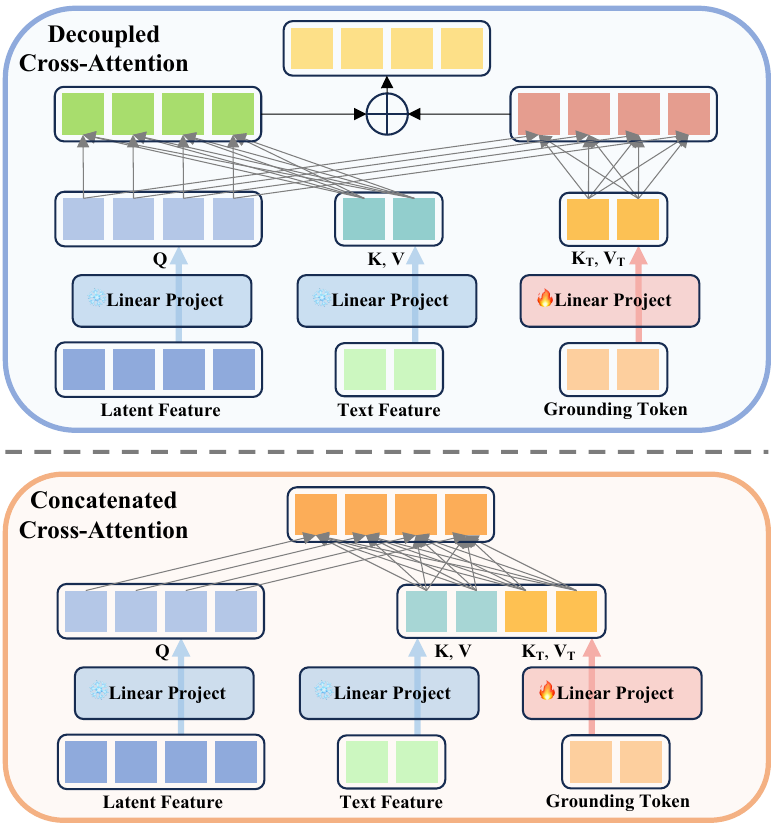}
   \vspace{-5mm}
   \caption{Comparison between decoupled cross-attention (DCA) and concatenated cross-attention (CCA) structures. DCA separately calculates two control attention maps and combines them using element-wise addition. In contrast, CCA integrates the two control conditions through concatenation, computing a single attention map that unifies layout and text control in one step.}
   \vspace{-5mm}
   \label{fig:CCA}
\end{figure}

\begin{figure*}
  \centering
  \includegraphics[width=1.0\linewidth]{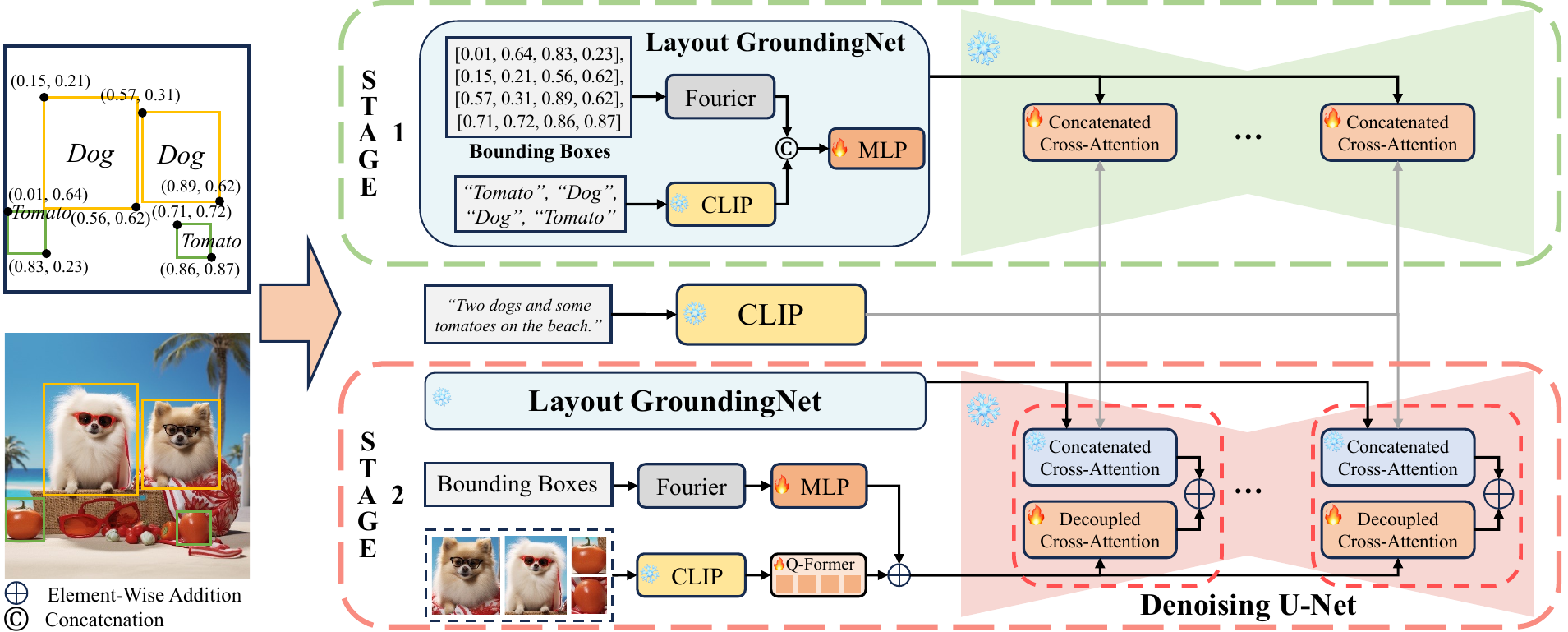}
  \vspace{-5.5mm}
  \caption{The proposed Progressive Two-Stage Training pipeline. In the first stage, we train the concatenated cross-attention network to incorporate layout control conditions, establishing an accurate layout-controllable generation model. In the second stage, we build on this pre-trained layout model by adding training for subject synthesis. This two-stage strategy results in an effective model for layout-controllable multi-subject synthesis.}
  \label{fig:model_framework}
  \vspace{-4.5mm}
\end{figure*}

\noindent \textbf{Explicit Layout Semantic Expansion.} To address control collision, we provide a simpler solution from a semantic perspective: 
For text-to-image tasks, the global semantics of the generated image should align with the global semantics of the corresponding text.  Subject synthesis information is also a form of global image semantics, and DCA effectively applies subject synthesis control information across the entire image through the element-wise addition in \cref{eq:decoupled cross attention} \cite{ye2023ip, wang2024instantid}, allowing it to align with the text control. However, layout is a type of implicit, local semantic information. Specifically, for the same text, the layout across multiple generated images is typically random, yet this randomness doesn’t affect image semantic alignment with the text. Consequently, when using DCA to apply explicit layout control information across the entire image, it may collide with the image’s implicit local layout semantics. A simple method to avoid this collision is to explicitly expand the implicit layout semantic information into the text, which represents the global semantics of the image.

We propose a semantic expansion method via concatenated cross-attention (CCA). As illustrated in \cref{fig:CCA}, compared to DCA, CCA also adds linear layers within the original CA layer to project \( G_T \) to key and value, denoted as \( K_T \) and \( V_T \). These are then concatenated with the key and value projections of the encoded text control, and the combined result is computed as follows:
\begin{equation}
\label{eq:concat cross attention}
\text{CCA} = \sigma\left(\frac{Q [K, K_T]^T}{\sqrt{d}}\right)[V, V_T].
\end{equation}
Unlike DCA’s dual computation, CCA considers both text and layout control in a single computation.
The purpose of CCA is to expand the text semantics to include explicit layout information. For example, let’s define a text prompt for generating an image as follows:  
\vspace{-1.5mm}
\begin{gather}
    ``A \enspace dog \enspace and \enspace a \enspace cat \enspace on \enspace the \enspace grass." \nonumber
\nonumber
\end{gather}
\vspace{-1.5mm}
 We want this text to be explicitly expanded to:
\begin{gather}
      \enspace ``A \enspace dog \enspace and \enspace a \enspace cat \enspace on \enspace the \enspace grass, \nonumber \\ \enspace with \enspace the \enspace cat \enspace located \enspace at \enspace B_{\text{cat}} \enspace and \enspace the \enspace dog \enspace at \enspace B_{\text{dog}}."
\nonumber
\end{gather}
 where \( B_{\text{cat}} \) and \( B_{\text{dog}} \) represent bounding box coordinates for the cat and dog.

\subsection{Decompose LMS into Progressive Stages.}
For the complete LMS setup, the grounding token \( G_I \) is obtained by replacing the subject text feature \( f_T^i \) in \cref{eq:grounding text} with the encoded subject image feature \( f_I^i \). Different methods \cite{li2023gligen,wang2024ms,zhou2024migc++} vary in the details of generating \( G_I \), and generally, both \( G_T \) and \( G_I \) are used to enhance performance.
Given DCA's effectiveness in handling subject synthesis, it remains a favorable choice for LMS. MS-Diffusion \cite{wang2024ms}  
achieves LMS via DCA. It performs well in subject synthesis but struggles with layout control due to the control collision issue discussed in \cref{subsec:explicit layout semantic expansion}. A straightforward solution involves constructing a composite network that injects \( G_T \) via CCA to achieve semantically controlled image layout, while also using DCA to inject \( G_I \) for subject synthesis in specified positions. 
The final calculation of CA in \cref{eq:cross attention} is modified as follows:
\begin{equation}
\label{eq:cross attention2}
\text{FCA}=\sigma\left(\frac{Q[K, K_T]^T}{\sqrt{d}}\right)[V, V_T] \\+ \lambda\cdot \sigma \left(\frac{Q K_I^T}{\sqrt{d}}\right)V_I.
\end{equation}
\noindent\textbf{Progressive Two-Stage Training.}
In practice, another challenge arises: training for layout control and subject synthesis simultaneously proves difficult, as they are complex and often conflicting sub-tasks. With the integration of CCA, the model’s layout controllability improves, but subject synthesis tends to suffer. To address this, we decompose the LMS task into a progressive two-stage task. Given the strong compatibility between CCA’s semantic expansion and text control, Stage 1 involves training a CCA-based layout control model. Once the model reaches satisfactory layout controllability, it is frozen, and in Stage 2, DCA is introduced to train the final LMS model. With layout controllability already integrated, Stage 2 training focuses on enhancing subject synthesis. The progressive two-stage training pipeline is depicted in \cref{fig:model_framework}.

\noindent \textbf{End-to-End Inference.}
With the progressive training strategy, the functionalities of both stages are successfully integrated into a single model, allowing for end-to-end, single-stage inference by simply inputting both \( G_T \) and \( G_I \). 

\noindent \textbf{Grounding Token Synthesis Improvements.}
To further enhance detail preservation in subject synthesis, we follow the IP-Adapter-Plus \cite{ye2023ip} by using a resampler with learnable query tokens. This resampler extracts 4 tokens representing subject image features from the 256 image feature tokens encoded by CLIP. However, using \cref{eq:grounding text} to fuse multiple tokens of the same subject image individually with the same bounding box disrupts the consistency of image information. We modify the method for obtaining \( G_T \) by projecting the encoded bounding box directly to match the image feature token dimensions using a separate MLP, and then directly adding it. The expression is as follows:
\begin{equation}
\label{eq: add grounding}
\text{$G_I^i$} = [\text{MLP}(\text{Fourier}(B_i)) + f_I^{i_j}],  \quad(1\leq j \leq 4)    
\end{equation}
where \( f_I^{i_j} \) represents the encoded features of the same subject image. After being fused with \( B_i \), they are concatenated to form the grounding token \( G_I \) for a single subject image.

\begin{figure*}
  \centering
  \includegraphics[width=1.0\linewidth]{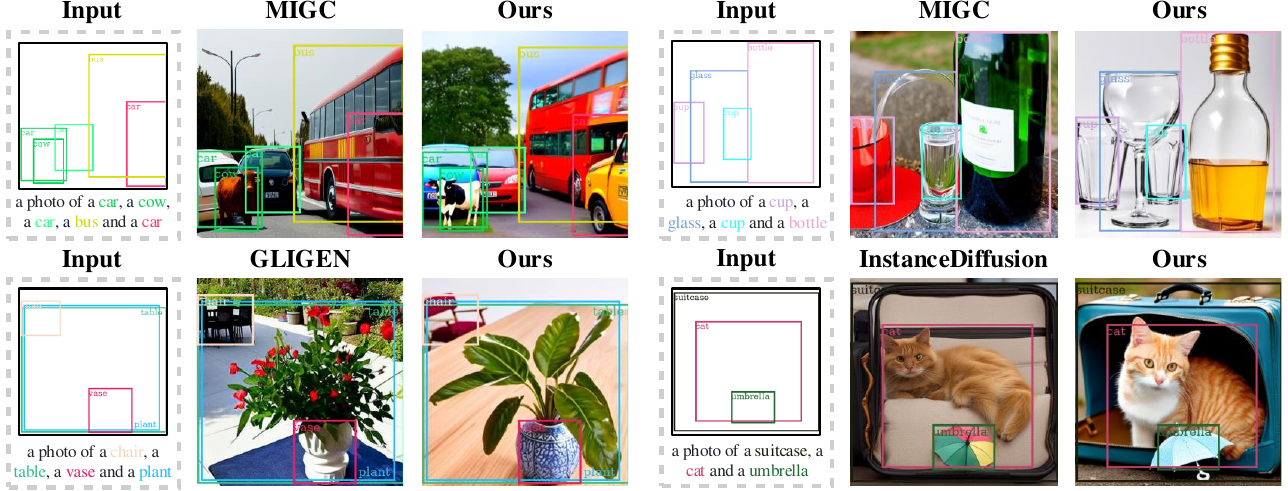}
  \vspace{-4.5mm}
  \caption{Qualitative experiments on MIG Bench.}
  \label{fig: mig-bench}
  \vspace{-2.5mm}
\end{figure*}

\section{Experiments}
\label{sec:experiments}
\subsection{Implementation Details}
\textbf{Training Data Preparation.}
We construct training data by randomly sampling 4M image-text pairs from the LAION-5B dataset \cite{schuhmann2022laion}. To enable precise subject localization, we employ a two-stage annotation process: 1) Subject labeling using the Recognize Anything Model \cite{zhang2024recognize}, followed by 2) Bounding box generation via Grounding DINO \cite{liu2023grounding}. This pipeline produces cropped subject images with text labels and accurate spatial coordinates.

\noindent\textbf{Experimental Setup.}
Our MUSE model is built on the pre-trained SDXL \cite{podell2023sdxl}, utilizing CLIP \cite{radford2021learning} as both the image and text encoder. For all training stages, we use the AdamW optimizer with a learning rate of \(1 \times 10^{-4}\). Training is conducted on 64 V100 GPUs, with a batch size of 384, for a total of 100K training steps. Two-stage resolutions of 512 and 1024 are used for training.
During inference, we employ the DDIM \cite{song2020denoising} sampler with 30 steps and set the CFG \cite{ho2022classifier} to 7.5, and for a fair comparison, 512 resolution is used. The DCA control scale $\lambda$ is empirically set to 0.8.

\subsection{Benchmark Datasets and Evaluation Metrics}
\textbf{MIG Bench.}
The MIG Bench dataset \cite{zhou2024migc}, constructed from 800 COCO \cite{lin2014microsoft} images, provides holistic text descriptions, instance-level color/class annotations, and bounding boxes for 2–6 instances per image. Its evaluation focuses solely on \textbf{layout success rate} (IoU\textgreater0.5 for all instances) through Grounding DINO \cite{liu2023grounding}-based detection. This design stems from two considerations: 1) In LMS tasks, instance colors are predefined by reference images rather than text descriptions, making color accuracy assessment irrelevant; 2) Layout fidelity constitutes the core challenge in multi-instance generation. Success rates are categorized by instance count (L2–L6) to analyze model scalability.
% \vspace{0.5em}

\begin{table}
\centering
% \vspace{-3mm}
\resizebox{0.48\textwidth}{!}{
\begin{tabular}{c|c c c c c |c|c}
% \hline
\toprule
% \rowcolor{gray!20}
\hline
 & \multicolumn{6}{c|}{Layout Success Rate (\%)} & \\
\cmidrule{2-7} 
        % \hline
        % \rowcolor{gray!20}
        \multirow{-2}{*}{Methods}& L2 & L3  &L4 &L5  & L6 & Avg &\multirow{-2}{*}{Time (s)}\\

        % \hline
        \hline
        GLIGEN    &0.913 & 0.908   & 0.877 & 0.834 & 0.848 & 0.866  &15.9\\
        % \hline
% \bottomrule
        InstanceDiffusion   &0.931  & 0.902 & \textbf{0.894} & 0.839 & 0.864 &0.876 & 26.5\\
        % \hline
        MIGC  &\textbf{0.934} & \textbf{0.925}   & 0.873 & 0.850 & 0.832 & 0.869  & 9.1\\
        % \hline
\bottomrule
        \textbf{Ours}   &0.906 & 0.917 & 0.891 & \textbf{0.858} & \textbf{0.878} & \textbf{0.884} &\textbf{4.1}\\
% \hline
\bottomrule
\end{tabular}
}
\vspace{-1.5mm}
\caption{Comparative evaluation of layout control performance on MIG Bench. Metrics include layout success rates (IoU\textgreater0.5) across instance count levels (L2–L6) and average inference time.}
\label{tab:migc}
\vspace{-5mm}
\end{table}

\noindent\textbf{MS-Bench.}
The MS-Bench dataset \cite{wang2024ms} contains 1,148 subject combinations (2–3 subjects per image) from 40 distinct classes, paired with bounding boxes and up to 6 text prompts per combination (4,488 total samples). To overcome the limitations of MS-Bench's rigid layout structure (primarily centered or left-right arrangements with uniform sizing), we propose \textbf{MS-Bench-Random}, which introduces randomized subject positions and sizes to better align with the practical requirements of LMS tasks. For evaluation, we measure: 1) \textbf{CLIP-T} score \cite{hessel2021clipscore} for text-image alignment; 2) \textbf{CLIP-I-local} score comparing cropped generated regions (using GT bounding boxes) with reference subjects via CLIP-I \cite{radford2021learning}; 3) \textbf{LMS success rate} requiring all subjects to meet CLIP-I-local thresholds (0.6/0.65). This multi-metric approach ensures a comprehensive assessment of spatial accuracy and subject fidelity.

\begin{figure*}
  \centering
  \includegraphics[width=0.98\linewidth]{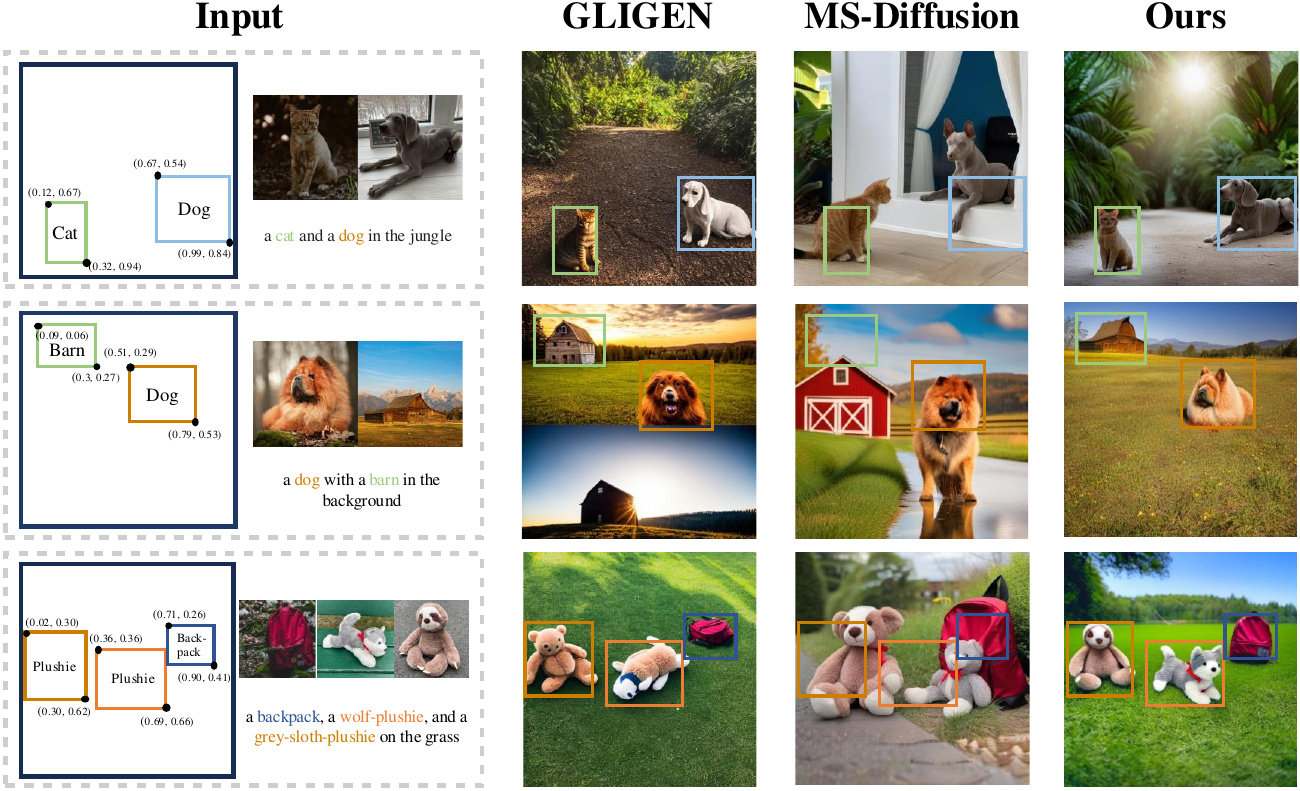}
  \vspace{-1.5mm}
  \caption{Qualitative experiments on MS-Bench-Random. }
  \label{fig:ms-bench}
  \vspace{-1.5mm}
\end{figure*}

% \vspace{2mm}
\begin{table*}
\centering
% \vspace{-3mm}
\resizebox{0.7\textwidth}{!}{
\begin{tabular}{c|c c c c | c c c c}
% \hline
\toprule
% \rowcolor{gray!20}
\hline
&  \multicolumn{4}{c|}{MS-Bench} &  \multicolumn{4}{c}{MS-Bench-Random} \\
\cmidrule{2-9} 
        % \hline
        % \rowcolor{gray!20}
        \multirow{-2}{*}{~~~~~~~~Methods~~~~~~~~}& ~CLIP-T~ & CLIP-I-l& ~SR-0.6~  &~SR-0.65~ &~CLIP-T~  & CLIP-I-l & ~SR-0.6~ & ~SR-0.65~ \\

        \hline
        % \hline
        GLIGEN   & 0.299 & 0.799   & 0.880 & 0.810 & 0.297 & 0.747  &0.870 & 0.682\\
        % \hline
        MS-Diffusion   & 0.318 & 0.780   & 0.820 & 0.706 & 0.316 & 0.686 &0.506 &0.289 \\
        \hline
\bottomrule
        \textbf{Ours}  & \bf{0.323} & \bf{0.827}  & \textbf{0.890} & \textbf{0.819} & \textbf{0.321} & \textbf{0.779} &\textbf{0.894} &\textbf{0.755}\\
% \hline
\bottomrule
\end{tabular}
}
\vspace{-1.5mm}
    \caption{LMS performance comparison on MS-Bench. Metrics include text-image alignment (CLIP-T), local subject fidelity (CLIP-I-l), and success rates using CLIP-I-l thresholds of 0.6/0.65 (SR-0.6/SR-0.65), evaluated under fixed and randomized layouts.}
\label{tab:ms-bench}
\vspace{-3.9mm}
\end{table*}

\begin{figure*}
  \centering
  \includegraphics[width=0.9\linewidth]{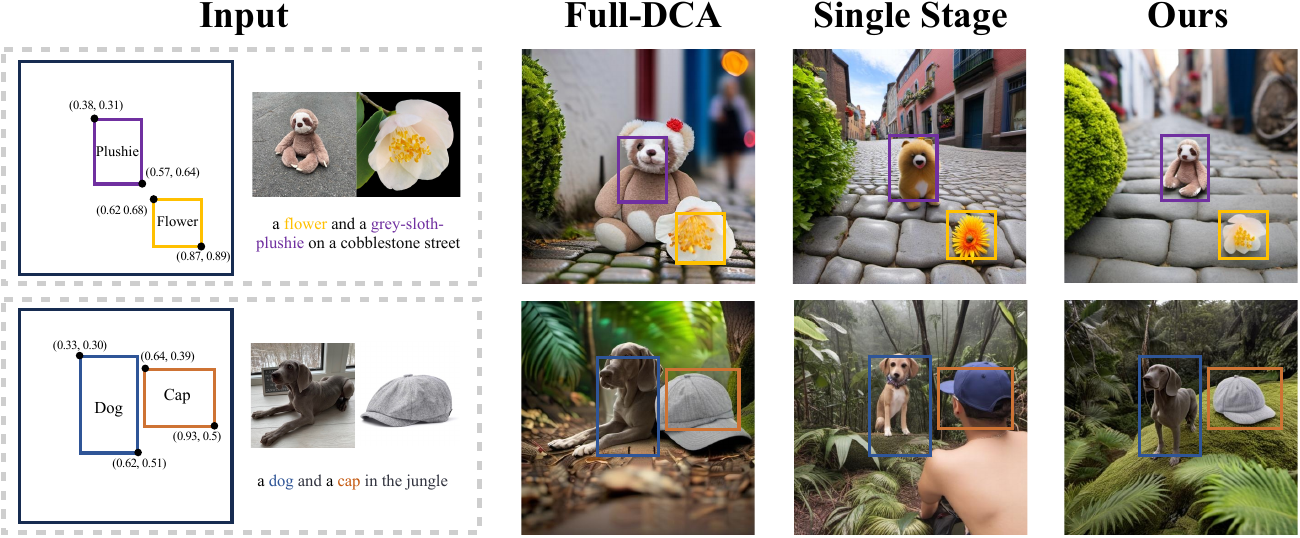}
  \vspace{-1.5mm}
  \caption{Ablation study on training strategies conducted on MS-Bench-Random. 
  }
  \label{fig:ab ms-bench}
  \vspace{-3.5mm}
\end{figure*}

\subsection{Quantitative Experiments}

\textbf{Layout Control on MIG Bench.}
As shown in \cref{tab:migc}, all compared methods achieve competent layout success rates (\textgreater86.6\% on average), demonstrating their effectiveness in addressing control collision issues through other innovations (\eg, GLIGEN's gated self-attention \cite{li2023gligen} or MIGC's divide-and-conquer approach \cite{zhou2024migc}). Our method establishes state-of-the-art performance with an 88.4\% average success rate. The advantage becomes particularly pronounced in high-complexity scenarios involving 5–6 instances (L5-L6). This superior scalability stems from our simple yet effective semantic expansion paradigm: The concatenated cross-attention (CCA) mechanism resolves control collisions through unified semantic alignment between layout constraints and textual descriptions, avoiding the need for complex module stacking or massive parameter growth. Notably, our lightweight implementation achieves inference speeds of 4.1s (3.9–6.5$\times$ faster than GLIGEN and InstanceDiffusion\cite{wang2024instancediffusion}), while maintaining architectural compatibility for downstream tasks through standardized attention interfaces, as evidenced by the seamless integration of subject synthesis.

% \vspace{-0.5mm}
\noindent\textbf{LMS Performance on MS-Bench.}
As Shown in \cref{tab:ms-bench}, our method demonstrates remarkable robustness under challenging randomized layouts. While baseline methods like MS-Diffusion \cite{wang2024ms} suffer catastrophic performance degradation in random conditions (SR-0.65 drops from 70.6\% to 28.9\%, -41.7\%), our approach maintains strong local subject fidelity (CLIP-I-local: 0.779) with only marginal quality reduction (SR-0.65: 75.5\%, -6.4\% from fixed layout). This minimal performance decay contrasts sharply with GLIGEN's 12.8\% SR-0.65 decline under equivalent randomization, confirming our MUSE framework effectively combines layout control with subject synthesis. Notably, even with randomized layout, our CLIP-T score (0.321) remains comparable to fixed layouts (0.323), indicating stable text-alignment capability regardless of spatial constraints.

\vspace{-1mm}
\subsection{Qualitative Experiments}
\textbf{Layout Control on MIG Bench.}
As shown in \cref{fig: mig-bench}, we compare our method with existing state-of-the-art methods \cite{zhou2024migc,wang2024instancediffusion,li2023gligen} in terms of layout control. It can be observed that our method is still able to accurately generate images and maintain good quality even in complex layout scenarios. This demonstrates the superiority of the CCA method in balancing layout control and text control.

\noindent\textbf{LMS Performance on MS-Bench-Random.}
\cref{fig:ms-bench} compares our MUSE model with state-of-the-art LMS methods \cite{li2023gligen,wang2024ms}, on the challenging MS-Bench-Random dataset. Our model consistently outperforms baselines in complex scenarios, achieving precise layout precision while preserving high-fidelity subject details. These results validate MUSE's effectiveness in real-world applications requiring both spatial accuracy and identity preservation.

\begin{table}
\centering
% \vspace{-3mm}
\resizebox{0.48\textwidth}{!}{
\begin{tabular}{c|c|c c c c c|c}
% \hline
\toprule
% \rowcolor{gray!20}
\hline
&  &\multicolumn{6}{c}{Layout Success Rate} \\
\cmidrule{3-8} 
        % \hline
        % \rowcolor{gray!20}
        \multirow{-2}{*}{Methods}&\multirow{-2}{*}{CLIP-T}& L2& L3&L4&L5&L6&Avg\\

        % \hline
        \hline
        DCA   & 0.322&0.807 & 0.801   & 0.780 & 0.761 & 0.756 & 0.781  \\
        % \hline
        CCA  & 0.331&0.906 & 0.917 & 0.891 & 0.858 & 0.878 & 0.884 \\
% \hline
\bottomrule
\end{tabular}
}

\vspace{-2mm}
\caption{Ablation study comparing attention mechanisms on layout control performance (MIG Bench). 
}
\label{tab:ablation cca}
\vspace{-5mm}
\end{table}

\subsection{Ablation Experiments}

\textbf{Explicit Layout Semantic Expansion Mechanism.} \cref{tab:ablation cca} reveals that replacing decoupled cross-attention (DCA) with our CCA framework improves layout success rates by 10.3\% (L2–L6 average) while maintaining superior text alignment (CLIP-T: 0.331 vs 0.322). This confirms CCA's ability to integrate layout constraints as expanded textual semantics rather than competing control signals.

\noindent \textbf{Progressive Two-Stage Training Strategy.}
\cref{ab tab:ms-bench training strategy} provides critical insights into the design of training strategies for our MUSE framework. The \textbf{full-DCA} approach reveals severe interference when applying DCA for both layout control and subject synthesis, resulting in a 41.0\% drop in SR-0.65. The \textbf{single-stage CCA+DCA} method reveals the inherent trade-off between concurrent layout and subject optimization: a reduced success rate loss (-33.6\% SR-0.65) comes at the expense of degraded subject fidelity (-0.085 CLIP-I-l). \textbf{Joint training}, which simultaneously optimizes both layout and subject objectives throughout the second training stage, achieves moderate success (-4.3\% SR-0.65). The \textbf{Reversed training} strategy, which prioritizes subject synthesis before layout control, proves catastrophic for spatial reasoning, as evidenced by a 0.056 CLIP-I-l gap (-7.2\%) compared to our approach. Our progressive two-stage strategy resolves these issues through ordered parameter freezing: CCA-first training establishes robust spatial priors, enabling subsequent DCA-based subject refinement in the second stage without control collision. 
The subject similarity of the \textbf{concatenated grounding} method in \cref{eq:grounding text} decreases (-0.015 CLIP-I-l while only -0.8\% SR-0.6), demonstrating our proposed grounding method in \cref{eq: add grounding} better maintains internal consistency among multiple image tokens of the subject.

\begin{table}
\centering
% \vspace{-3mm}
\resizebox{0.48\textwidth}{!}{
\begin{tabular}{c|c c c c }
% \hline
\toprule
% \rowcolor{gray!20}
\hline
\cmidrule{2-5} 
        % \hline
        % \rowcolor{gray!20}
        ~~~~~~~~~Methods~~~~~~~~& ~CLIP-T~ & CLIP-I-l& ~SR-0.6~  &~SR-0.65~  \\

        \hline
        Full-DCA & -0.013 & -0.076 & -0.309 & -0.410 \\
        Single Stage & -0.008 & -0.085 & -0.227 & -0.336 \\
        Joint Training & -0.005 & -0.022 & -0.017 & -0.043 \\
        Reversed Training & -0.008 & -0.056 & -0.043 & -0.084 \\
        Concatenated Grounding & -0.003 &-0.015 &-0.008 &-0.021 \\
\bottomrule
        Ours  & 0.321 & 0.779 &0.894 &0.755\\
% \hline
\bottomrule
\end{tabular}
}
\vspace{-2.5mm}
    \caption{Ablation study on training strategies under randomized layouts(MS-Bench-Random). 
    }
\label{ab tab:ms-bench training strategy}
\vspace{-5mm}
\end{table}

\cref{fig:ab ms-bench} qualitatively evaluates our progressive two-stage LMS framework against full-DCA and single-stage CCA+DCA methods on MS-Bench-Random. Our approach uniquely achieves precise layout control and high-fidelity subject synthesis, demonstrating superior suitability for LMS tasks compared to competing training strategies.

\vspace{-1mm}
\section{Conclusion}
\label{sec:conclusion}
In this research, we introduce MUSE, a novel framework addressing LMS task. Our method decomposes the complex LMS task into a progressive, two-stage sub-task. We first introduce the CCA method, which enhances the model's compatibility between layout and text control by explicitly expanding the layout semantic for text, effectively resolving control collision. Building upon this enhanced layout control model, we employ the DCA method to enable multi-subject synthesis, achieving zero-shot LMS without modifying pre-trained models. Extensive experiments validate the effectiveness of MUSE, demonstrating its superior performance on LMS tasks and valuable contribution to enhancing control ability in text-to-image diffusion models.
{
    \small
    \bibliographystyle{ieeenat_fullname}
    \bibliography{main}
}

\clearpage
\setcounter{page}{1}
\maketitlesupplementary

In this supplementary material, we provide more design details of our layout-controllable multi-subject synthesis (LMS) method MUSE, along with extensive experimental results. These include comparisons between our proposed MUSE and other LMS methods, evaluations of concatenated cross-attention (CCA) versus decoupled cross-attention (DCA), ablation studies on the progressive two-stage training strategy, and ablation experiments on the subject synthesis strength scale. Additionally, in the \textbf{$MUSE\_AttnProcessor.py$} file, we provide the implementation of the final cross-attention operation that integrates both the CCA and DCA methods. This implementation is built using the diffusers  \cite{von-platen-etal-2022-diffusers} library.

\section*{A. More Details of Encoding Control Information}
For the text for layout control and the images for subject synthesis, we use CLIP-ViT-L-14 and CLIP-ViT-G-14 \cite{radford2021learning} models for encoding, respectively. 

For layout text features, we extract the class token from the CLIP model's final output. For a class text like ``dog", the encoded feature has a size of [1,768]. Bounding box information is Fourier-encoded with a frequency of 16, producing a feature of size [1, 64]. These two features are concatenated along the feature dimension, resulting in a [1, 768+64] feature, which is further processed by an MLP to produce a fused grounding token of size [1,768]. The MLP consists of three linear layers with SiLU activation functions.

For image encoding, the straightforward approach is to use the same final class token output of CLIP encoding (size [1, 1280]) concatenated with bounding box features to create the grounding token. While sufficient for simple text class, this is inadequate for subject synthesis, as the class token lacks rich spatial information. Existing subject synthesis works like IP-Adapter \cite{ye2023ip}, RealCustom \cite{huang2024realcustom}, and InstantID \cite{wang2024instantid} utilize shallower CLIP features, such as the last hidden states (size [256, 1664]), which offer ample spatial information. However, excessive spatial detail leads to redundancy in subject synthesis (\eg, copy-paste artifacts), especially since SDXL \cite{podell2023sdxl} itself uses only 77 tokens for text prompts control.

To address this, we adopt IP-Adapter-Plus's approach: train 4 learnable tokens as queries, processed through four layers of perceiver attention, extracting compressed image features of size [4, 2048]. Directly concatenating bounding box information ([4, 2048+64]) into the tokens for MLP fusion degrades subject synthesis, as independently processed tokens may produce inconsistent results due to neural network black-box behavior. Instead, we independently encode the Fourier-transformed bounding box information into [1, 2048] through an MLP, and add this layout encoding to each image token, achieving coherent grounding tokens. Mapping the dimension of the image grounding token to 2048 is intended to initialize the mapping layer parameters in DCA using those from the pretrained model, thereby reducing the training difficulty of the model.

\section*{B. More Details of Experimental Setup}
Since the data samples used in both training and testing contain multiple subjects, we set the number of subjects per sample to 10. For samples with more than 10 subjects, we select the 10 largest bounding box areas. For samples with fewer than 10 subjects, we use padding by introducing trainable empty tokens to maintain 10 subjects. These include text, image, and coordinate features. During training, we randomly drop captions for images and conditions (\eg, subject texts, images and bounding boxes) for MUSE with a 0.1 probability to enhance robustness.

Regarding the loss function, we maintain consistency with the original pre-trained model. The diffusion network is trained to accurately predict added noise under additional multiple control conditions.

For each test experiment, we use five random seeds and report the average results.

\section*{C. More Qualitative Experiment Results} 

\begin{figure*}
  \centering
  \includegraphics[width=1.0\linewidth]{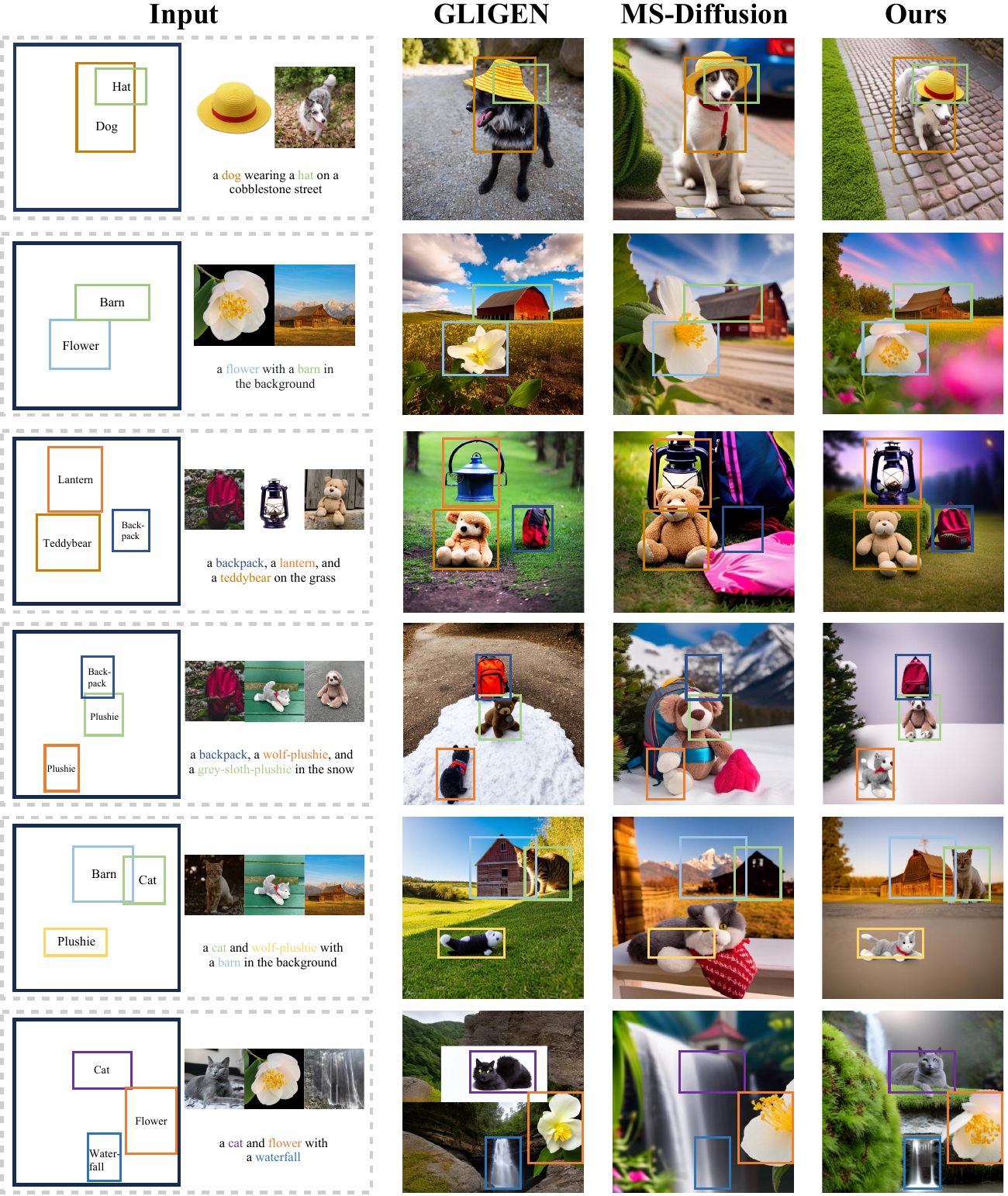}
  % \vspace{-1.5mm}
  \caption{Qualitative experiments on MS-Bench-Random. Our method demonstrates strong LMS performance across various conditions, showing good practical applicability in real-world.}
  \label{sup fig: ms-bench}
  % \vspace{-2mm}
\end{figure*}

\cref{sup fig: ms-bench} provides more qualitative comparisons of our method against other LMS approaches, including GLIGEN \cite{li2023gligen} and MS-Diffusion \cite{wang2024ms}, on the MS-Bench-Random dataset. Our approach consistently demonstrates superior performance in both layout control and multi-subject synthesis.

\begin{figure*}
  \centering
  \includegraphics[width=1.0\linewidth]{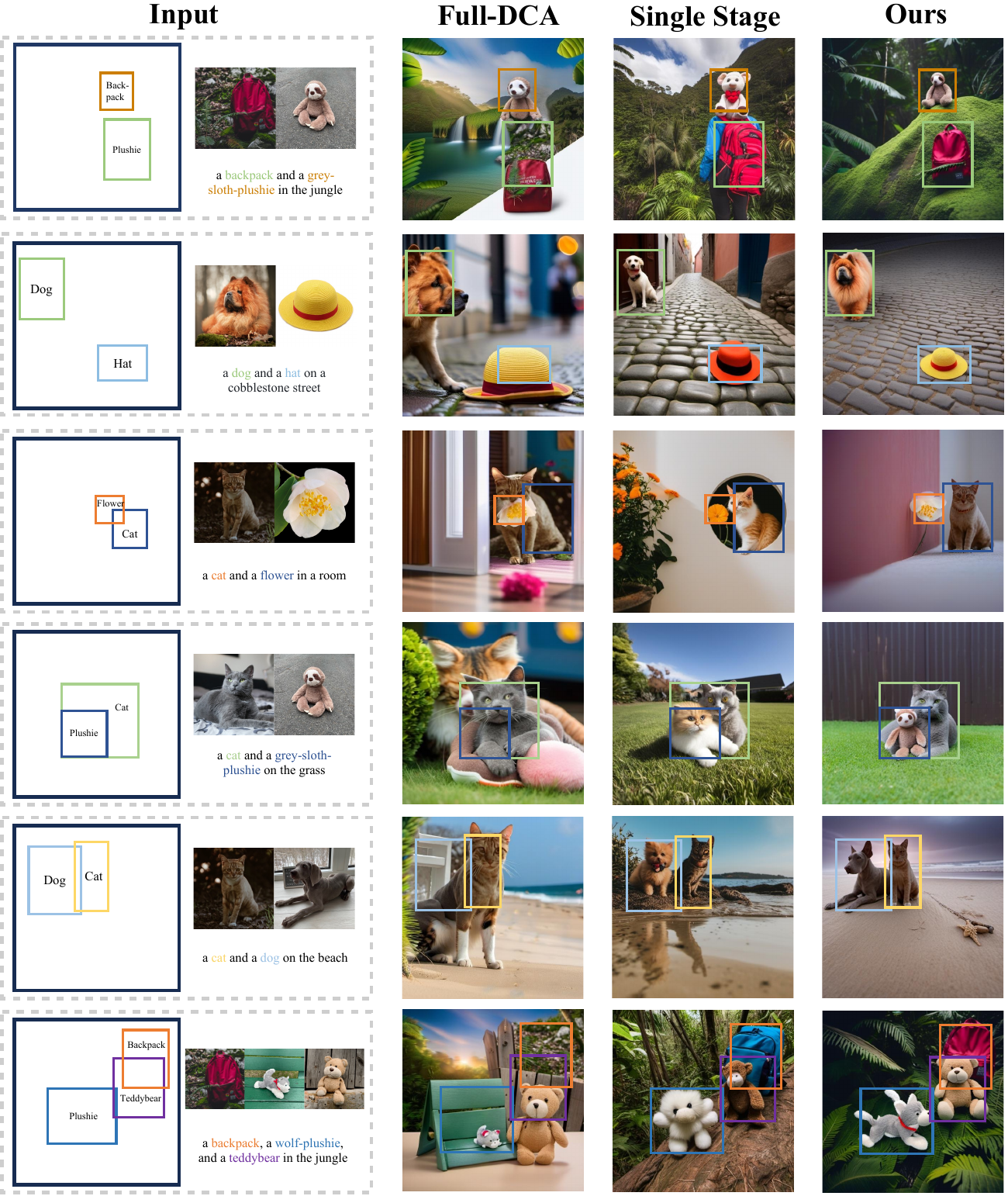}
  % \vspace{-1.5mm}
  \caption{Ablation study on training strategies conducted on MS-Bench-Random. It shows that using both CCA and DCA methods along with our proposed progressive two-stage training strategy significantly improves performance on the LMS task.}
  \label{sup fig:ab ms-bench} 
  % \vspace{-2mm}
\end{figure*}

\begin{table*}
\centering
% \vspace{-3mm}
\resizebox{0.95\textwidth}{!}{
\begin{tabular}{c|c c c c | c c c c}
% \hline
\toprule
% \rowcolor{gray!20}
\hline
&  \multicolumn{4}{c|}{MS-Bench} &  \multicolumn{4}{c}{MS-Bench-Random} \\
\cmidrule{2-9} 
        % \hline
        % \rowcolor{gray!20}
        \multirow{-2}{*}{~~~Scale~~~}& ~CLIP-T~ & CLIP-I-l& ~SR-0.6~  &~SR-0.65~ &~CLIP-T~  & CLIP-I-l & ~SR-0.6~ & ~SR-0.65~ \\

        \hline
        % \hline
        0.6   & 0.328 & 0.811   & 0.878 & 0.790 & 0.327 & 0.769  &0.871 & 0.712\\
        \hline
        0.8   & 0.323 & 0.827 & 0.890 & 0.819 & 0.321 & 0.779 &0.894 &0.755\\
        \hline
\bottomrule
        1.0  & 0.313 & 0.832   & 0.889 & 0.822 & 0.310 & 0.782 &0.897 &0.763 \\
% \hline
\bottomrule
\end{tabular}
}
% \vspace{-1.5mm}
    \caption{Ablation experiments on the subject synthesis strength scale. The evaluated metrics include CLIP-T, CLIP-I-local (abbreviated as CLIP-I-l), and LMS Success Rate (SR), determined using CLIP-I-local score thresholds of 0.6 and 0.65, referred to as SR-0.6 and SR-0.65, respectively.}
\label{sup tab:ms-bench scale}
% \vspace{-3mm}
\end{table*}

\section*{D. More Ablation experiments on Training Strategies}

\cref{sup fig:ab ms-bench} presents qualitative results on the MS-Bench-Random dataset, comparing our progressive two-stage LMS framework with models trained using the full-DCA method and single-stage training combining CCA and DCA method. Our proposed framework achieves both accurate layout control and effective subject synthesis. The results confirm that our proposed progressive framework is more suitable for LMS tasks.

\section*{E. Ablation experiments on Subject synthesis Strength Scale}

We conducted ablation experiments on the subject synthesis strength scale \( \lambda \). While the default sacle is set to 0.8, we provide results for \( \lambda = 0.6 \) and \( \lambda = 1.0 \) on the MS-Bench-Random dataset. Quantitative comparisons are shown in \cref{sup tab:ms-bench scale}. 

The results indicate that high subject synthesis strength scale can weaken text-following ability, while low strength scale reduces subject synthesis quality. A balanced value achieves optimal performance.

\end{document}